\documentclass[final,3p]{elsarticle}

\usepackage{lmodern}
\pdfoutput=1
\usepackage{booktabs}

\usepackage[T1]{fontenc}
\usepackage{textcomp}
\usepackage{algorithm} 
\usepackage{xcolor}

\usepackage{acronym}

\usepackage{enumitem}

\setlist[itemize]{parsep=0pt, topsep=0.1\baselineskip, itemsep=0.2\baselineskip}
\setlist[itemize,1]{leftmargin=2\parindent, label=\textbullet}
\setlist[itemize,2]{leftmargin=\parindent, label=--}

\setlist[enumerate]{parsep=0pt, topsep=0.1\baselineskip, itemsep=0.2\baselineskip}
\setlist[enumerate,1]{leftmargin=2\parindent, label=\textup{\arabic*.}, ref=\arabic*}
\setlist[enumerate,2]{leftmargin=\parindent, label=\textup{\alph*)}, ref=\theenumi.\alph*}

{\end{itemize}\end{minipage}}

\usepackage{graphicx}
\usepackage[caption=false,font=footnotesize]{subfig}
\usepackage{dblfloatfix}
\usepackage{array}
\usepackage{multirow}
\usepackage{longtable}

\usepackage[breakable,skins,raster]{tcolorbox}

\definecolor{ColTblRed}{RGB}{255,16,57}\definecolor{ColTblGrn}{RGB}{0,194,157}
\definecolor{ColTblBlu}{RGB}{54,135,213}\definecolor{ColTblOrn}{RGB}{244,148,85}
\definecolor{ColTblYlw}{RGB}{237,192,33}\definecolor{ColTblMgn}{RGB}{164,123,237}
\definecolor{ColTblCyn}{RGB}{73,183,218}\colorlet{ColTblDflt}{ColTblCyn}
\newtcolorbox{TblBox}[4][]{enhanced, center, width=#3, before skip=6pt, halign=flush center, colframe=\ifx\\#1\\#2\else#1\fi!75!black, colback=#2!10, coltitle=white, boxrule=1pt, boxsep=1.5pt, segmentation style={solid, line width=0.5pt}, toptitle=4.5pt, bottomtitle=3pt, top=4pt, left=9pt, right=9pt, bottom=9pt, halign title=flush center, fonttitle=\small\bfseries, fontupper=\footnotesize, fontlower=\footnotesize, title={#4}}

\newcounter{AlgCnt}

\usepackage[noend]{algpseudocode} 
\definecolor{ColDrkGry}{gray}{0.41}
\colorlet{ColAlgLinNum}{ColDrkGry}
\colorlet{ColAlgCom}{ColDrkGry}
\definecolor{ColAlgBck}{RGB}{241,237,228}
\definecolor{ColAlgFrm}{RGB}{35,37,49}
\algrenewcommand{\alglinenumber}[1]{\color{ColAlgLinNum}\footnotesize#1:}
\algrenewcommand{\algorithmiccomment}[1]{\hfill\textcolor{ColAlgCom}{$\triangleright$\textit{#1}}}
\algnewcommand{\AlgLngCmnt}[1]{\Comment{\parbox[t]{.67\linewidth}{#1\strut}}}
\algnewcommand{\LongComment}[1]{\Comment{\parbox[t]{.67\linewidth}{#1}}}

\usepackage[cmex10]{amsmath}
\usepackage{amssymb}
\usepackage{amsthm}
\usepackage{mathrsfs}
\usepackage{units}
\allowdisplaybreaks

\usepackage{hyperref}
\definecolor{ColHypBlu}{RGB}{33,113,181}
\hypersetup{colorlinks=true,allcolors=ColHypBlu}

\definecolor{ColAuthCmnt}{Hsb}{0,1,0.8}
\newcommand{\AuthCmnt}[2][]{\textcolor{ColAuthCmnt}{\textsf{\textless\ifx\\#1\\\textbf{\boldmath #2}\else\textbf{@#1: }#2\fi\textgreater}}}
\definecolor{ColAuthChng}{Hsb}{210,0.95,0.75}

\journal{\AuthCmnt{Elsevier Journal}}

\makeatletter
\def\hyp{\leavevmode\nobreak\hskip\z@skip-\discretionary{}{}{}\nobreak\hskip\z@skip}
\def\nobrkhyp{\leavevmode\nobreak\hskip\z@skip\hbox{-}\nobreak\hskip\z@skip}
\makeatother

\begin{document}

\setlength{\abovedisplayskip}{0.75ex plus2pt minus1pt}
\setlength{\belowdisplayskip}{0.75ex plus2pt minus1pt}
\setlength{\abovedisplayshortskip}{0pt plus 2pt}
\setlength{\belowdisplayshortskip}{0.75ex plus2pt minus1pt}
\begin{frontmatter}


\title{A Novel Strategy for Improving Robustness in Computer Vision Manufacturing Defect Detection}

\author[add:1,add:2]{Ahmad~Mohamad~Mezher\corref{cor:1}}
\ead{ahmad.mezher@unb.ca}

\author[add:1]{Andrew~E.~Marble}
\ead{andrew@willows.ai}

\cortext[cor:1]{Corresponding author. Tel.: \AuthCmnt{+1 000 000 0000}.}

\address[add:1]{Department of Electrical and Computer Engineering, University of New Brunswick, (\href{http://www.unb.ca/}{UNB}), Fredericton, NB, Canada}

\address[add:2]{Willows AI, (\href{https://www.willows.ai/}{Willows AI Inc.}), Montreal, QC, Canada}


\begin{abstract}
Visual quality inspection in high performance  manufacturing can benefit from automation, due to cost savings and improved rigor. Deep learning techniques are the current state of the art for generic computer vision tasks like classification and object detection. Manufacturing data can pose a challenge for deep learning because data is highly repetitive and there are few images of defects or deviations to learn from. 
Deep learning models trained with such data can be fragile and  sensitive to context, and can under-detect new defects not found in  the training data. In this work, we explore training defect detection models to learn specific defects out of context, so that  they are more likely to be detected in new situations. We  demonstrate how models trained on diverse images containing a common defect type can pick defects out in new circumstances. Such  generic models could be more robust to new defects not found data collected for training, and can reduce data collection impediments to implementing visual inspection on production lines. Additionally, we demonstrate that object detection models trained to predict a label and bounding box outperform classifiers that predict a label only on held out test data typical of manufacturing inspection tasks.
Finally, we studied the factors that affect generalization in order to train models that work under a wider range of conditions.


\textcopyright\ 2023 The Authors. Preprint submitted to Elsevier, Inc.
\end{abstract}


\end{frontmatter}

\setcounter{footnote}{0}
\renewcommand{\thefootnote}{(\arabic{footnote})}


\section{Introduction}
\label{sec:1}
\noindent
In the manufacture of mechanical products in complex industrial processes, defects such as internal holes~\cite{PENG201769}, pits~\cite{C5TB00270B}, abrasions~\cite{7855792}, underfill~\cite{Oleff2021} , and scratches~\cite{CHEN2015106} arise, due to failures in design, production equipment and production environment conditions. Products may also easily corrode~\cite{Rodionova2016} and be prone to fatigue because of daily use. These defects increase the costs incurred by enterprises, including warranty and reputation costs, shorten the service life of manufactured products, and result in an extensive waste of resources, and may cause substantial harm to people and their safety~\cite{Wang2018}. Hence, detecting defects is a core competency that enterprises should possess in order to improve the quality of the manufactured products without affecting production~\cite{Wang2018,LIAO201963,Kim2018}. Automatic defect-detection technology has obvious advantages over manual detection~\cite{Park2016}. It not only adapts to an unsuitable environment but also works in the long run with high precision and efficiency, and does not suffer from fatigue or variability as with a human inspection. Research on defect-detection technology can reduce the production cost, improve production efficiency and product quality, as well as lay a solid foundation for the intelligent transformation of the manufacturing industry~\cite{HUANG20151}.

 Supervised machine learning inspection methods, including classification and object  detection, usually depend on access to a set of annotated images  for training. This presents a problem in manufacturing, where  defect rates are low, making it difficult or impossible to collect  a representative set of images of the defects that are likely to be  encountered~\cite{15158,9000806}. Additionally because production line data  generally consists of repeated images of nearly identical parts,  there is a high risk of over-fitting to available data, resulting in a  model that may perform well in training and validation, but is  not robust to changes that can occur in production~\cite{10.1145/3368555.3384468}. The  overall result is models that can fail inexplicably under different  conditions.
 
 Anomaly detection has been used to overcome some drawbacks of supervised learning, by training models to learn “OK” parts and flag any that deviate. This approach is most suitable for highly uniform images, but can become more challenging for parts with significant natural variation. In this case, many images are still required, small defects can be lost within natural variability,~\cite{s20071829} and there is no guarantee that the model is learning a set of features that adequately represent the defects.
 
 Deep learning is often lauded for its generalization ability: models learn concepts rather than specific rules, and so can work on unseen examples. This is true for example in classifiers trained on the over 14 million images in Image-Net, that can predict class labels for subjects in new situations~\cite{https://doi.org/10.48550/arxiv.1409.1556}. Generalization is achieved by exposing the model to a diverse variety of examples, incentivizing it to learn broad concepts of what differentiates, say, a bird from a plane, rather than looking for rules.
 
 For many domain specific data sets, including manufacturing inspection, the model does not see a diverse set of examples. For defect inspection, the background image is often the same (the part itself) and defects may frequently be of the same type or in the same location. Under these circumstances, models are more apt to learn shortcuts that may not generalize to new defects or even causally relate to the presence of a defect in the image~\cite{https://doi.org/10.48550/arxiv.2003.08907}. Such shortcuts can make the model less robust to variations in the data encountered in production~\cite{10.1145/3368555.3384468}.

This work addresses the over-fitting problem in defect inspection by training a ML model on a data set that contains diverse external data, featuring defect types that are of interest in a variety of contexts. We experimentally vary the background object for a class of defects and examine how classification and object detection performance compares with a training set containing near identical objects as would be encountered in a typical manufacturing inspection scenario.


\subsection{Contribution and Organization}
\label{sec:CAO}
\noindent
Our contributions are summarized as follows:

\begin{itemize}
    \item We have created data sets and training  models that incentivize learning at the concept level, by varying  the background characteristics of the image.
	\item We have shown that an object model trained on diverse data that includes defect instances in different contexts and materials can generalize to defects in new situations.
	\item We have validated our approach using several experiments that show that a generalization effect. 
    \item We have shown that object detection models provide better performance compared to classifiers in terms of area under the receiver operator characteristic curve (AUC) when generalizing to new backgrounds.
    \item We have used a clustering method to study the factors that affect generalization in order to train models that work under a wider range of conditions.  
\end{itemize}

The paper is structured as follows. Relevant related works are presented in Section \ref{sec:RW}. Defect detection based on machine learning methods, including unsupervised learning, traditional supervised learning and deep learning, are reviewed in Section \ref{sec:back}. Performance metrics used for evaluating defect detection models are provided in Section \ref{sec:metrics}. A summary of data augmentation is explained in Section \ref{sec:DA}. Section \ref{sec:ExS} explains the data collection  and labeling and the model selected for experiments. Simulation results are provided in Sections \ref{sec:res1} and \ref{sec:tl}. Section \ref{sec:res2} identifies the key factors that can affect the generalization of models. Finally, conclusions and future works are given in Section \ref{sec:Conclusion}.

\section{Related work}
\label{sec:RW}
\noindent

Efforts to improve generalization can largely be grouped into three categories: out-of-distribution (OOD) detection, synthetic  data generation, and additional data collection. OOD detection  flags model inputs that are outside the training data distribution, for example by looking at disagreement over an ensemble of models~\cite{https://doi.org/10.48550/arxiv.1612.01474} or at the distance of an input from the training data in a modified feature space~\cite{https://doi.org/10.48550/arxiv.2102.11582}. Anomaly detection can be considered equivalent to OOD flagging for this discussion. While a promising and valuable component of a visual inspection system, OOD detection methods can add complexity, require sufficient data to model the distribution well enough to minimize false positives, and require that the model is responsive to new anomaly features in order to register them as outside the distribution. For example, a model that learns to ignore a region or features in an image because they do not contribute to minimizing loss during training may not capture information from features from that region should a defect occur there, and so not “see” the image as OOD.

Synthetic data generation includes programmatic image transformations that change the data distribution~\cite{9710159}, alterations such as moving or pasting elements of other classes into images~\cite{https://doi.org/10.48550/arxiv.2007.09438}, and synthesis such as by style transfer~\cite{https://doi.org/10.48550/arxiv.1910.03334} or GANs~\cite{15158,9000806}. The limitation of these methods is that they are constrained by the variability of the available data and any manually added variation, and therefore may not be able to fully capture real differences that arise in production data. Additionally, generalization from synthetic to real data presents additional challenges and is not guaranteed~\cite{15158}.

Adding more data can improve performance and generalization. Ref.~\cite{Bukhsh2021} combined six separate data sets showing images of damaged concrete, and showed that models trained through transfer learning on the combined set had better overall performance. They did not evaluate generalization outside the six data sets. A challenge with this approach for many defect types is the limited number of data sets, the lack of diversity within each data set. 

\section{Problem description}
\label{sec:back}
\noindent
Computer vision is regularly proposed to inspect manufactured products, and in particular neural network models are often recommended for their ability to generalize to defects beyond the the examples encountered in training. In theory, Convolutional Neural Network (CNN) models have demonstrated generalization power. For example, classifiers and object detection models trained on publicly available data sets (such as Image-Net and COCO [https://cocodataset.org/] respectively) can perform their task on new instances that differ from the training data. Classifiers trained on Image-Net have been used to identify new images that are not part of that data set.

In practice, the generalization power of CV models trained on large research data sets is rarely extended to domain specific applications like manufacturing. While data sets like Image-Net have millions of different images and considerable variation, manufacturing data sets typically have (regardless of number) images of more-or-less the same component over and over again. Also, instances of defects or variations are rare for mature manufacturing processes, and can under-represent the total range of defects that could emerge. If we use such data that is highly homogeneous, class imbalanced, and has few examples of defects to train a machine learning model, the result may be highly over-fit to the data that is available. Such an over-fit model will not provide any of the generalization advantage of a model trained on diverse data, and in fact may be just as brittle, or more so, than a rules based classical computer vision model. The reason it could be worse is that trained models can find shortcuts that don’t actually relate to any causal features of the image, whereas a model based on human-derived rules and some classical image manipulation is at least tuned to look at defects.

A corollary to the above situation is the difficulty in validating an inspection model. With only an imbalanced, homogeneous data set available, it is challenging to convincingly demonstrate the ability of trained models to correctly identify new defects.

Conceptually, one can imagine that the success of training on big data sets like Image-Net could be transposed to inspection problems: a model that has been trained on thousands of images of different defects, for example, would be expected to be incentivized to learn some general features of a defect (rather than just memorize a shortcut) and be able to identify a defect in some new material, size, and orientation, when presented with one. Such a detector has big advantages in manufacturing, because it does not require training data to begin working, and has a lower risk of misidentifying a “new” defect, so long as it’s still (in this example) a defect.

Our approach is to show that an object model trained on diverse data that includes defect instances in different contexts and materials can generalize to defects in new situations, and that by using transfer learning, it is possible to train a performant model using only a few hundred instances. To make such models practical and reliable, we want to investigate how well they generalize, and how generalization can be improved, so that a user of the models can confirm they meet inspection standards and will catch defects that arise. This approach can be applied with any machine learning methods used to detect surface defects in manufactured goods.

In the next subsection, we will describe several machine learning methods used to detect manufacturing defects.

\subsection{Relevant machine learning methods}
\label{sec:dl}
\noindent

Researchers have developed multiple deep learning models to identify defects in industrial manufacturing. This section focuses on machine learning algorithms that are specifically used for defect detection, which are categorized into two subsections. The first subsection discusses defect detection methods based on classification, while the second focuses on methods based on object detection models.

\subsubsection{Defect detection method based on classification}
\label{sec:ddc}
\noindent


Defect classification can predict the presence of a defect in an image and label the image accordingly. Classification can be binary - OK or not-good (NG) or multi-class if more labels are needed. The support vector machine (SVM)~\cite{Suykens1999} and K nearest neighbor (KNN)~\cite{6313426} are two well-known classifiers that have been commonly applied to defect inspection.


Support vector machines (SVM) are a popular machine learning tool that are suitable for small and medium-sized data samples, as well as for nonlinear, high-dimensional classification problems. They have been extensively used in the industrial vision detection field. For instance, a real-time machine vision system that uses SVMs to learn complex defect patterns was proposed by the authors in~\cite{1334512}. In~\cite{LI2009374}, a binary defect pattern classification method that combines a supervised SVM classifier with unsupervised self-organizing map clustering was proposed, in which SVMs are employed to classify and identify manufacturing defects. The method achieved over 90\% classification accuracy, which outperformed the back-propagation neural network. However, this study only focused on binary map classification. For multi-class defect detection and classification, ~\cite{VALAVANIS20107606} proposed a method based on a multi-class SVM and a neural network classifier for weld radiographs. ~\cite{VALAVANIS20107606} established an improved SVM classification model based on a genetic algorithm for real-time analysis of spectrum data to accurately estimate different types of porosity defects in an aluminum alloy welding process. Moreover, SVM classifiers have played a significant role in inspecting surface defects in copper strips~\cite{10.1504/IJCAT.2012.045840}, monitoring and diagnosing defects in laser welding processes~\cite{6803929}, defect detection in wheel bearings~\cite{1542014}, and more.


The KNN algorithm has demonstrated greater simplicity and stability compared to neural networks~\cite{10.1016/j.patcog.2005.08.009, LEI20091535}. In~\cite{doi:10.1177/1528083714555777}, the authors utilized a sequence of pre-processing techniques, including wavelet, threshold, and pathological operations, to prepare images for defect detection. They then employed the grey-level co-occurrence matrix (GLCM) method to extract features before using the KNN algorithm to classify defect images. The overall accuracy rate of this classification approach was around 96\%.



The authors of~\cite{7878212} proposed a method that utilizes multiple image texture feature extraction techniques. They combined local binary pattern (LBP) with the grey level run length matrix (GLRLM) to extract image features and employed KNN and SVM for classification. The experimental results indicated that combining LBP and GLRLM can improve feature extraction performance, and SVM outperforms nearest neighbor methods for texture feature classification. Alternatively, an unsupervised algorithm can be applied for defect classification. A multi-objective fault signal diagnosis problem can be solved efficiently using a genetic algorithm-based method that relies on K-means clustering~\cite{10.1145/3230905.3230952}. Additionally,~\cite{8316611} introduces an unsupervised defect detection algorithm for patterned fabrics. This algorithm divides a filtered image into a series of blocks and inputs the squared difference between each block median and the mean of all block medians into K-means clustering to classify the blocks as OK or NG. The overall detection success rate was found to reach 95\%.


Industrial production has greatly benefited from recent advances in neural networks, generally considered to be encompassed under the term Artificial Intelligence (AI). In particular, deep learning, which uses an increased number of network layers, has become the standard in supervised computer vision. Deep learning is able to automatically learn and extract features with strong predictive value for computer vision tasks, and can reduce the amount of feature engineering and fine-tuning required.


The convolutional neural network (CNN) is the most widely used architecture for classifying images. LeNet's emergence in 1998 marked the beginning of CNNs~\cite{ 726791}. In 2012, AlexNet's success in the Image-Net competition popularized deep learning in computer vision, and numerous CNN models have since emerged, including Network-in-Network~\cite{https://doi.org/10.48550/arxiv.1312.4400}, VGGNet~\cite{https://doi.org/10.48550/arxiv.1409.1556}, GoogLeNet~\cite{7298594}, ResNet~\cite{7780459}, and DenseNet~\cite{8099726}. A CNN consists of three primary types of neural layers that perform distinct functions: convolutional layers that identify local feature combinations from the previous layer, pooling layers that consolidate semantically similar features, and fully connected layers that ultimately transform feature maps into a feature~\cite{WANG2018144,https://doi.org/10.48550/arxiv.1807.08596}.

The CNN was initially designed for image analysis, making it suitable for automated defect classification in visual inspection~\cite{Du2021}. In recent years, deep learning has been applied to industrial defect classification in various fields, including industrial production and electronic components. For supervised steel defect classification, a max-pooling CNN approach was proposed in~\cite{6252468}. The CNN outperformed SVM classifiers and functioned correctly with different types of defects. Surface quality affects product appearance and performance. In~\cite{Park2016}, a generic CNN-based approach for automatic visual inspection of dirt, scratches, burrs, and wears on part surfaces was presented. Pre-trained CNN models achieved improved accuracy on small data sets for a surface quality visual inspection system. A robust detection method based on a visual attention mechanism and feature-mapping deep learning was proposed in~\cite{Lin2018} to detect casting defects by X-ray inspection. A CNN extracted defect features from potentially defective regions to obtain a deep learning feature vector, and the similarity of suspicious defective regions was calculated using the feature vector. The method effectively solved the problem of false and missing inspections. A CNN-based inspection system was proposed in~\cite{Nguyen2021} to achieve defect classification in casting products, but the CNN deep learning model performed well only with a large volume of high-quality data. In~\cite{Kim2020}, authors proposed an indicator to differentiate between defects and the background area for the classification of defect types in thin-film-transistor liquid–crystal display panels. For industrial production processes, automatic defect classification was performed based on a CNN.





Transfer learning is a technique in machine learning that involves leveraging a pre-existing model in a different task. This method can address the issue of limited labeled data. To illustrate, a CNN-based transfer learning approach for automatic defect classification was suggested in~\cite{8839832} where it was demonstrated that the technique is practical even with small training data sets, achieving over 80\% accuracy with just a few dozen labeled data points. For our study, we employ Image-Net as a pre-training data set, as previous research~\cite{7780459} has shown that ResNet-50, trained on Image-Net, serves as a reliable generic feature extractor and a suitable starting point for training.


\subsubsection{Defect detection method based on object detection models}
 \label{sec:ddm}
\noindent


Detecting objects in images is a crucial aspect of computer vision, which involves locating objects in an image using bounding boxes and determining their type. Object detection using deep learning methods can be broadly grouped into two categories. The first category generates regions and then classifies them to obtain various object categories, while the second category treats object detection as a regression or classification problem and uses a unified framework to directly obtain the final categories and locations~\cite{8627998}. Examples of region proposal-based methods include R-CNN~\cite{6909475}, spatial pyramid pooling (SPP-net)~\cite{7005506}, Fast R-CNN~\cite{7410526}, Faster R-CNN~\cite{7485869}, region-based fully convolutional networks (R-FCNs)~\cite{https://doi.org/10.48550/arxiv.1605.06409}, feature pyramid networks (FPNs)~\cite{8099589}, and Mask R-CNN~\cite{8237584}. Examples of regression-and classification-based methods include MultiBox~\cite{6909673}, AttentionNet~\cite{7410662}, G-CNN~\cite{7780629}, You Only Look Once (YOLO)~\cite{7780460}, the single-shot MultiBox detector (SSD)~\cite{10.1007/978-3-319-46448-0_2}, YOLOv2~\cite{8100173}, RetinaNet~\cite{8237586}, YOLOv3~\cite{https://doi.org/10.48550/arxiv.1804.02767}, and YOLOv4~\cite{https://doi.org/10.48550/arxiv.2004.10934}. Generally, region proposal-based methods have higher accuracy but are slower, while regression-and classification-based methods are faster but have lower accuracy.


A two-stage fabric defect detector based on a cascaded mixed feature pyramid network (FPN) was proposed by the authors in~\cite{s20030871}. They introduced a feature extraction backbone model that matches parameters with fitting degrees to address issues related to small defect feature space and background noise. Stacked feature pyramid networks were established to integrate cross-scale defect patterns for feature fusion and enhancement in a neck module. Moreover, they proposed cascaded guided region proposal networks (RPNs) to refine the anchor centers and shapes used for anchor generation. The experimental results demonstrated that this method can enhance the recognition performance across different scales.


Faster R-CNN is a cutting-edge technique for real-time object detection that uses an RPN to generate ROIs instead of selective search. For instance, the authors of~\cite{LEI2019379} proposed a Faster R-CNN method to perform intelligent fault detection for high voltage lines. The method selects a random region as the proposal region and then determines the corresponding category and location of a specific component after training. The experiments showed that the detection method, based on the ResNet-101 network model, was effective in identifying insulator damage and bird nests on a high voltage line. In~\cite{electronics8050481}, authors introduced an enhanced Faster R-CNN method for surface defect recognition in wheel hubs. They replaced the last maximum pooling layer with an ROI pooling layer that enabled the use of a single feature map for all the proposals generated by the RPN in a single pass. This technology allowed object detection networks to use an input feature map with a flexible size and output a fixed-size feature map. The experimental results demonstrated that the improved Faster R-CNN method achieved higher detection accuracy, at the expense of detection speed.


The object detection and recognition algorithm, ``You Only Look Once'' (YOLO), uses a deep neural network and fixed-grid regression to perform its functions quickly and is desigend for use in real-time applications~\cite{Wang2021}. Its unique feature is that it takes the entire image as input and directly determines the object's location and category at multiple positions in the image through regression. In~\cite{electronics9091547}, the YOLO/CNN model was employed by authors to detect defects on printed circuit boards (PCBs) and achieved a defect detection accuracy of 98.79\%. However, the types of defects that can be detected by this method are limited and require optimization. The authors in~\cite{s20061650} proposed an active learning method for steel surface defect inspection using YOLOv2. Results from extensive experiments on a challenging public benchmark demonstrate that the proposed approach is highly effective.





The SSD algorithm is a hybrid of YOLO and Faster R-CNN that employs multi-scale regional features for regression. This approach maintains the high speed of YOLO while ensuring a certain level of accuracy. For example, in~\cite{8978787}, a DF-SSD object detection method based on DenseNet and feature fusion was proposed to replace VGG-16 in SSD. A fusion mechanism for multiscale feature layers was also designed to effectively integrate low-level visual features and high-level semantic features. The experimental results indicated that the proposed DF-SSD method could achieve advanced performance in the detection of small objects and objects with specific relationships. However, for this work, we will be using RetinaNet, which is an SSD variant described in detail in Section~
\ref{sec:detectron2}.

\section{Performance metrics}
\label{sec:metrics}
\noindent


Selecting the right metrics is key to evaluating defect inspection models, as different metrics may prioritize different outcomes, such as the prevalence of different failure modes. In the following section, we will outline some commonly used performance evaluation metrics in the defect detection field.

\subsection{Confusion matrix}




In an inspection system designed to identify defective parts, we assume two possible labels for a part: it can either be deemed ``OK'' or ``NG'' (not good). An image is labelled as NG if it contains one or more defects. When the inspection system makes a prediction about the status of a part, it can either predict that the part is OK or predict that it is NG. 
To evaluate the accuracy of the inspection system, we can tabulate the image-level prediction results on a test set as in Table~\ref{tab:confus_mat} that shows the possible outcomes of the system's predictions. The table has four quadrants, with the predicted status of the image on one axis and the actual (``True'') status of the image on the other.

The top left quadrant represents true positives (TP), which occur when the inspection system correctly identifies an OK image as OK. The bottom right quadrant represents true negatives (TN), which occur when the system correctly identifies an NG image as NG.
The top right quadrant represents false negatives (FN), which occur when the system incorrectly identifies an OK image as NG. This can result in unnecessary costs or delays if the part is removed from the production line when it is actually acceptable.
The bottom left quadrant represents false positives (FP), which occur when the system incorrectly identifies an NG image as OK. An automated inspection system will trade off throughput (minimizing false positives that require manual inspection) with the chance of shipping a defective part (a false negative). The latter is usually considered a more serious error because it means that a defective part may go unnoticed and end up in the final product or shipped to a customer. However the false positive rate will generally determine whether automating inspection can be economically viable, because it this rate will determine how much manual work is avoided.


%

\begin{table}[h]
\centering
\begin{tabular}{c|cc}
\multicolumn{1}{c}{} & \multicolumn{2}{c}{\textbf{Predicted}} \\ 
\cline{2-3}
\textbf{True} & \textbf{OK} & \textbf{NG} \\
\hline
\textbf{OK} & True Positive (TP) & False Negative (FN) \\
\textbf{NG} & False Positive (FP) & True Negative (TN) \\
\end{tabular}
\caption{Confusion matrix for a binary classification problem}
\label{tab:confus_mat}
\end{table}

\subsection{ROC, AUC, IOU, and AP}

\begin{figure}[H]
	\centering
	\includegraphics[width=5in]{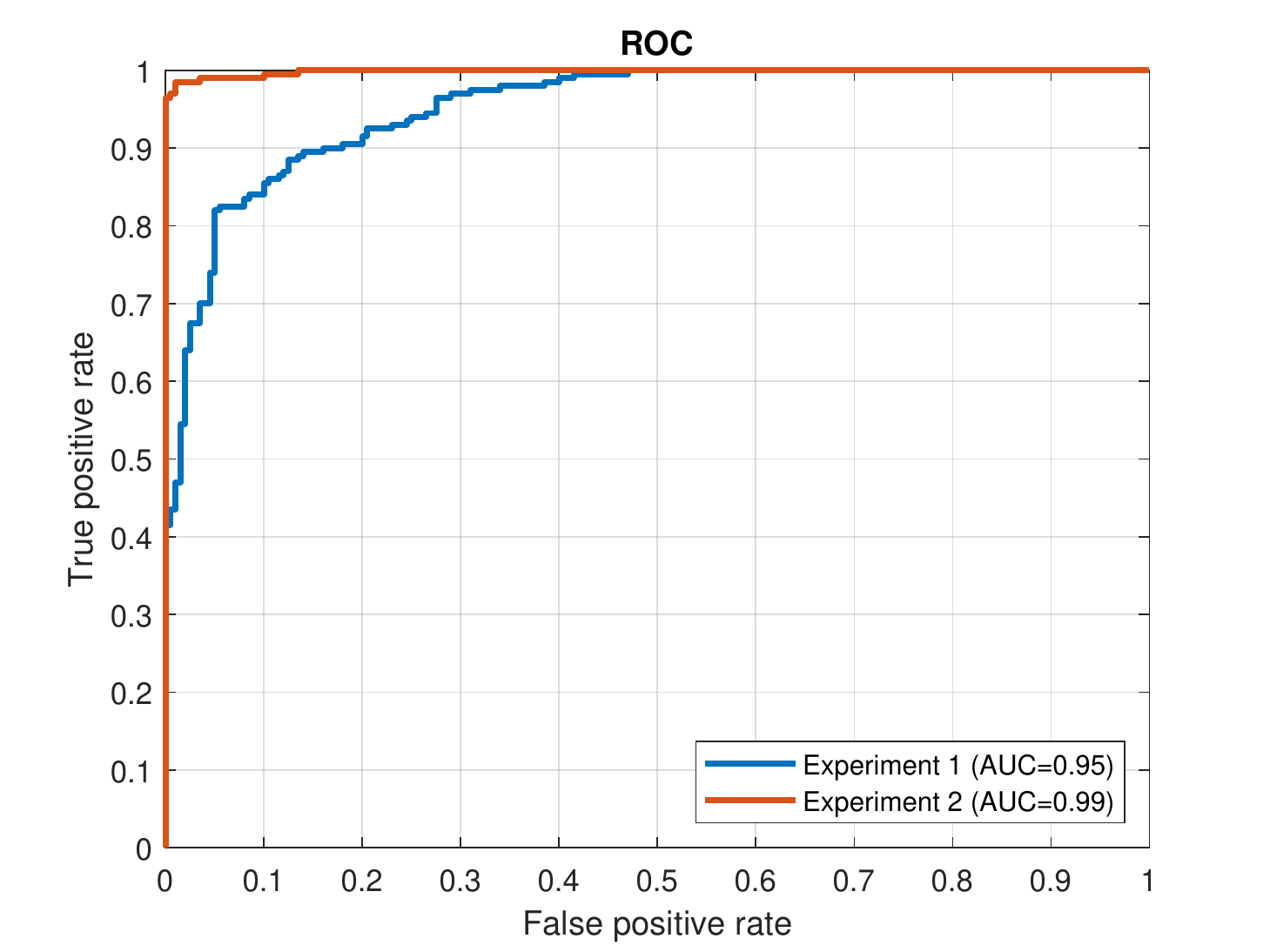}
	\caption{Two ROC curves for two different models.}
	\label{fig:AUC}
\end{figure}

For a model that outputs a score, assumed to be between 0 and 1 and representing predicted probability that there is a defect in the image, the labels in Table~\ref{tab:confus_mat} depend on the threshold selected to map a score to OK or NG. To evaluate detection performance, the ROC (Receiver Operating Characteristic) curve~\cite{doi:10.1177/0272989X8900900307} and AUC (Area Under Curve)~\cite{BRADLEY19971145} can also be used. The ROC curve plots the relationship between true positive (TP) rate and false positive (FP) rate, as demonstrated in Fig. \ref{fig:AUC}, which displays two ROC curves. Generally, when the ROC curve is closer to a step function, as in Experiment 2, the model is deemed to perform better, with low FP and high TP rates. The AUC, which is the area under the ROC curve, is often used to compare two ROC curves from different models.

When employing models like SSD for object detection, IOU (Intersection over Union) is frequently used to determine if an object is accurately localized. IOU measures the overlap rate between the bounding box provided by the model and the ground truth bounding box. If the IOU exceeds a predefined threshold, the object detection is considered successful. Object detection models for natural scenes often use scores of .25, .5, or .75 as thresholds to indicate success. For manufacturing inspection, it is more important that a defect be found than accurately bounded. Furthermore, some defect types, such as cracks, may not lend themselves well to being measured by IOU due to their aspect ratio and possibly subjective extent. 

\begin{equation}
      \label{eq:IOU}
      IOU= \frac{\text{Detection} \: \text{Result} \cap \text{Ground} \:  \text{Truth} }{\text{Detection} \:   \text{Result} \cup \text{Ground} \:  \text{Truth}}
    \end{equation}
    

The average precision (AP) metric used in object detection algorithms is based on the precision-recall curve, which shows the trade-off between the precision and recall of the detections for different threshold values. AP measures the area under this curve and provides a summary of the accuracy of the detections for all possible threshold values.

To compute the AP, we first calculate the precision and recall for each threshold value of the confidence score (i.e., the score assigned to each detected object by the algorithm). We then interpolate the precision values at each recall level to obtain a smooth curve and compute the area under this curve.

The equation for AP can be written as:
\begin{equation}
\text{AP} = \frac{1}{|\mathcal{D}|}\sum_{d=1}^{|\mathcal{D}|} \int_0^1 p_d(r) \mathrm{d}r
\end{equation}

where $\mathcal{D}$ is the set of test images, $d$ indexes the images in $\mathcal{D}$, $p_d(r)$ is the precision at recall level $r$ for image $d$, and $|\mathcal{D}|$ is the number of images in the test set.

The precision at recall level $r$ for image $d$ can be computed as:

\begin{equation}
p_d(r) = \max_{\tilde{r} \geq r} p_d(\tilde{r})
\end{equation}

where $p_d(\tilde{r})$ is the precision at recall level $\tilde{r}$ for image $d$.

In practice, the AP is often computed for a range of threshold values and averaged over all the test images to obtain a single value that summarizes the overall performance of the algorithm.

While AP is a useful metric for evaluating the overall performance of object detection algorithms, it may not always be the most appropriate metric for inspection tasks where minimizing false negatives is critical. This is because AP penalizes false negatives (i.e., missed detections) more severely than false positives (i.e., incorrect detections), which may not be desirable in some inspection scenarios. Additionally, AP can be sensitive to IOU as discussed which is less relevant for flagging defects.

In this work, we map object detection scores to (AUC,ROC) scores by assigning image level probabilities equal to the lowest probability predicted for any bounding box by the object detection model. This allows us to compare object performance with that of a classifier, and provides a more suitable score for evaluating inspection performance.

\subsection{Data sets for metric evaluation}

A common practice for reporting metrics is to divide a base data set in to two or three subsets termed \texttt{train}, \texttt{validation}, and optionally \texttt{test} or \texttt{holdout}. The model is trained on the \texttt{train} set and evaluated on the \texttt{validation} set that was not used in training. Optionally, the \texttt{holdout} set is used to confirm performance of a model that has been through multiple fine-tuning sessions on the \texttt{train} and \texttt{validation} data, making sure the model has not been inadvertently over-fit to the \texttt{validation} set. The model would be considered over-fit if the \texttt{holdout} performance is lower than on the other sets. 

Unless otherwise specified, it is usually assumed that the subsets are randomly split from some initial pool of data. Such evaluation implies that the distribution of the data is the same in all two or three subsets, and that there will not be ``new'' or ``unseen'' data in the \texttt{validation} or \texttt{holdout} spits. This differs from real cases where data may drift subtly or a long tailed distribution may result in never-before-seen defects in production. As such, performance on equal splits, even with a test split included, may not be indicative of actual production performance.

\section{Data Augmentation }
\label{sec:DA}
\noindent

Training a deep neural network model typically requires a significant amount of data, which can be expensive and time-consuming to acquire and label. Data augmentation is a useful technique that addresses this challenge by substantially increasing the diversity of available training data without the need for additional data collection. Common data augmentation methods include image rotation, flipping, mirroring, noise addition, and illumination alteration. These techniques are often combined to generate even more varied data.
 




\section{Experiments}
\label{sec:ExS}
\noindent

The aim of our experiments was to compare object detection and classification models' performance at identifying manufacturing defects in new contexts and on newly collected test sets, as a measure of the generality of the model. We additionally studied the role of training data, comparing a ``uniform'' data set consisting of near identical parts and backgrounds, which is typical of manufacturing, with a data set containing similar defects across a wide variety of parts. 

\subsection{Model selection}
\label{sec:detectron2}
\noindent

\subsubsection{ResNet Backbone}
\label{sec:ResnetBB}
\noindent

A ResNet-50 backbone was used in this study, configured as a binary classifier for classification experiments and as part of the RetinaNet object detetion model as described below. ResNet-50 is a commonly used architecture in machine learning computer vision tasks, and consists of repeated convolutional layers short circuited with identity operations, originally formulated to improve ease of training for very deep networks~\cite{7780459}. The use of this specific classifier was based on its proven success in image classification tasks, and the binary classifier head allowed for the specific classification needs of this study.

For the object detection model, we employed the RetinaNet SSD, based on the Detectron2 library~\cite{wu2019detectron2}. This library is Facebook AI Research's advanced software that offers cutting-edge segmentation and detection algorithms. All models used a RetinaNet model pre-trained on COCO as the starting point for training, with all model parameters allowed to train. The succeeding subsection will present a concise overview of the RetinaNet detector.

\subsubsection{RetinaNet detector}
\label{sec:RD}
\noindent


RetinaNet integrates the strengths of several target recognition approaches, most notably the "anchor" concept from RPN and the usage of feature pyramids from Single Shot Multibox Detector (SSD)~\cite{Liu_2016} and Feature Pyramid Networks (FPN)~\cite{https://doi.org/10.48550/arxiv.1612.03144}. The RetinaNet model consists of three components: a convolutional neural network for feature extraction and two sub-networks for classification and box regression~\cite{8417976}. Figure 9 illustrates the model structure, with Figure 9a depicting the ResNet-50 backbone network, Figure 9b showing how FPN serves as a decoder to generate a multi-scale convolutional feature pyramid, and Figure 9c revealing how two sub-networks are employed for classification and bounding box regression. Using feature mapping, the classification and box regression sub-networks are constructed via straightforward convolutional operations. The classification sub-network is responsible for object classification, while the box regression sub-network returns the bounding box position. FPN's advantage is that it leverages the hierarchical structure of the deep convolutional network to represent multi-scale objects, enabling the recognizer to create more accurate position predictions.

\begin{figure}[H]
	\centering
	\includegraphics[width=5in]{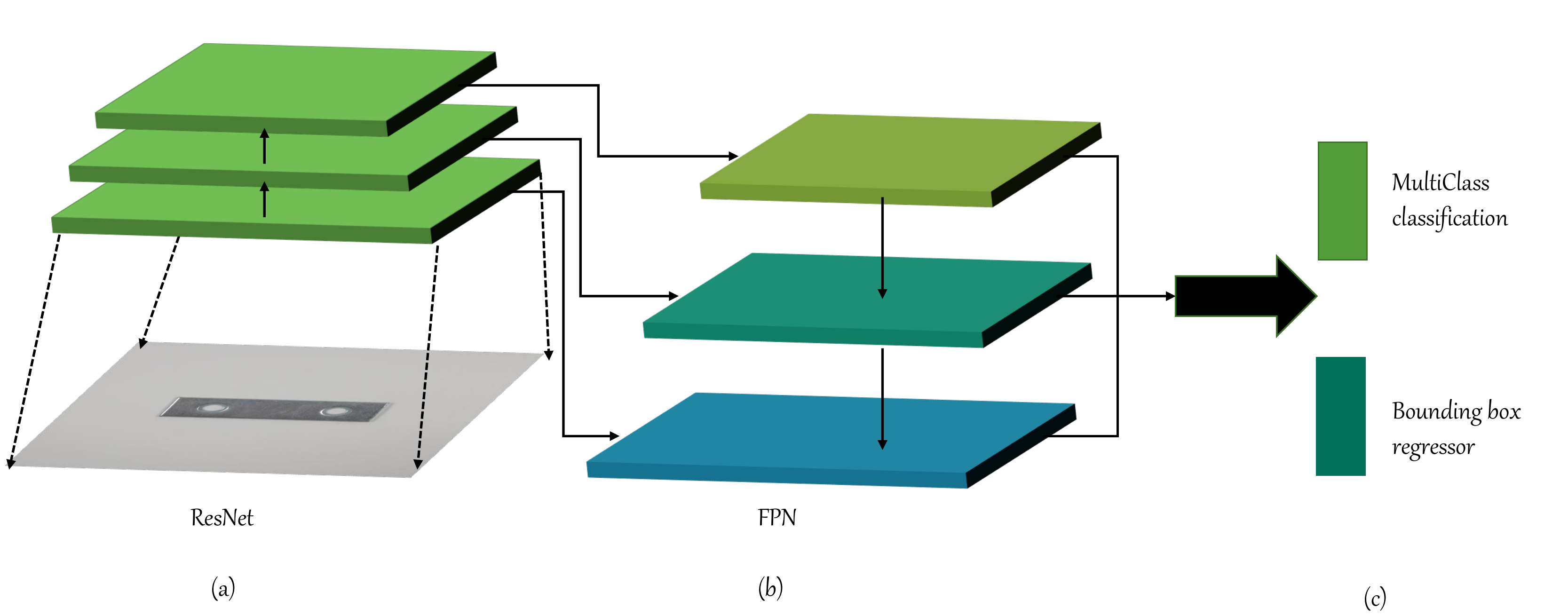}
	\caption{The structure of RetinaNet.(a) Backbone network; (b) decoder; (c) subnet.}
	\label{fig:micro}
\end{figure}


In this paper, ResNet-50 is utilized to extract image features~\cite{8417976}. Unlike the two-stage recognition method, the low accuracy of the one-stage target recognition is primarily due to the extreme imbalance between foreground and background samples during the training of the dense recognizer. This leads to an abundance of negative samples during the training process. To address this issue of category imbalance, focus loss is employed. This method modifies the standard cross-entropy by reducing the loss assigned to well-classified examples~\cite{8417976}. With focus loss supervision, RetinaNet is able to achieve remarkable improvements in the universal object recognition benchmark. Equation (\ref{eq:fl}) represents the focus loss and has been utilized to enhance detection accuracy. An a-balanced variation of the focus loss is defined as follows:

\begin{equation}
      \label{eq:fl}
      FL(p_t)= \alpha_{t}(1-p_t)^\gamma \log (p_t)
    \end{equation}


The hyperparameters $\alpha_{t}$ and $\gamma$ are used where $\alpha_{t}$ $\in$ [0,1] is the weight assigned to address class imbalance, and the parameter $\gamma$ adjusts the rate at which easy examples are down weighted. To simplify notation, $p_t$ is defined as:

\begin{equation}
p_{t} =
\begin{cases}
\text{p if y=1}.
\\
 \text{1-p otherwise}.
\end{cases}
\label{eq:alphastrategy}
\end{equation}

where $p$ $\in$ [0,1] is the probability estimated by the model, and $y = 1$ specifies the ground truth.



\subsection{Experimental setup}
\label{sec:Eset}
\noindent

The classifier experiments in this study used the torchvision~\cite{torchvision} implementation of a ResNet-50 classifier, pre-trained on Image-Net. During training, the classifier model was configured with the following parameter settings: Epoch: 200, learning rate: 0.0001, batch size per image: 16, and image size: 448.

To conduct object detection experiments, we used RetinaNet model from the Detectron2 library~\cite{wu2019detectron2}. The model used a ResNet-50 backbone pre-trained on COCO. The object detection model was fine tuned with the following parameter settings: Epoch: 9000, learning rate: 0.00025, and batch size per image: 128. All experiments were conducted on a single K80 or V100 GPU.


\subsection{Data collection and Labeling}
\label{sec:dcandl}
\noindent

A first data set consisting of photographs of 200 substantially identical metal “Mending plates” was assembled. Half of these parts were damaged by removing a roughly crescent shape portion of material using a 3~mm diameter round file, to simulate missing material caused by a manufacturing defect. Fig. \ref{fig:Mendingplates1} below shows 25 example images, showing the full structure of the part being inspected. 

\begin{figure}[H]
	\centering
	\includegraphics[width=5in]{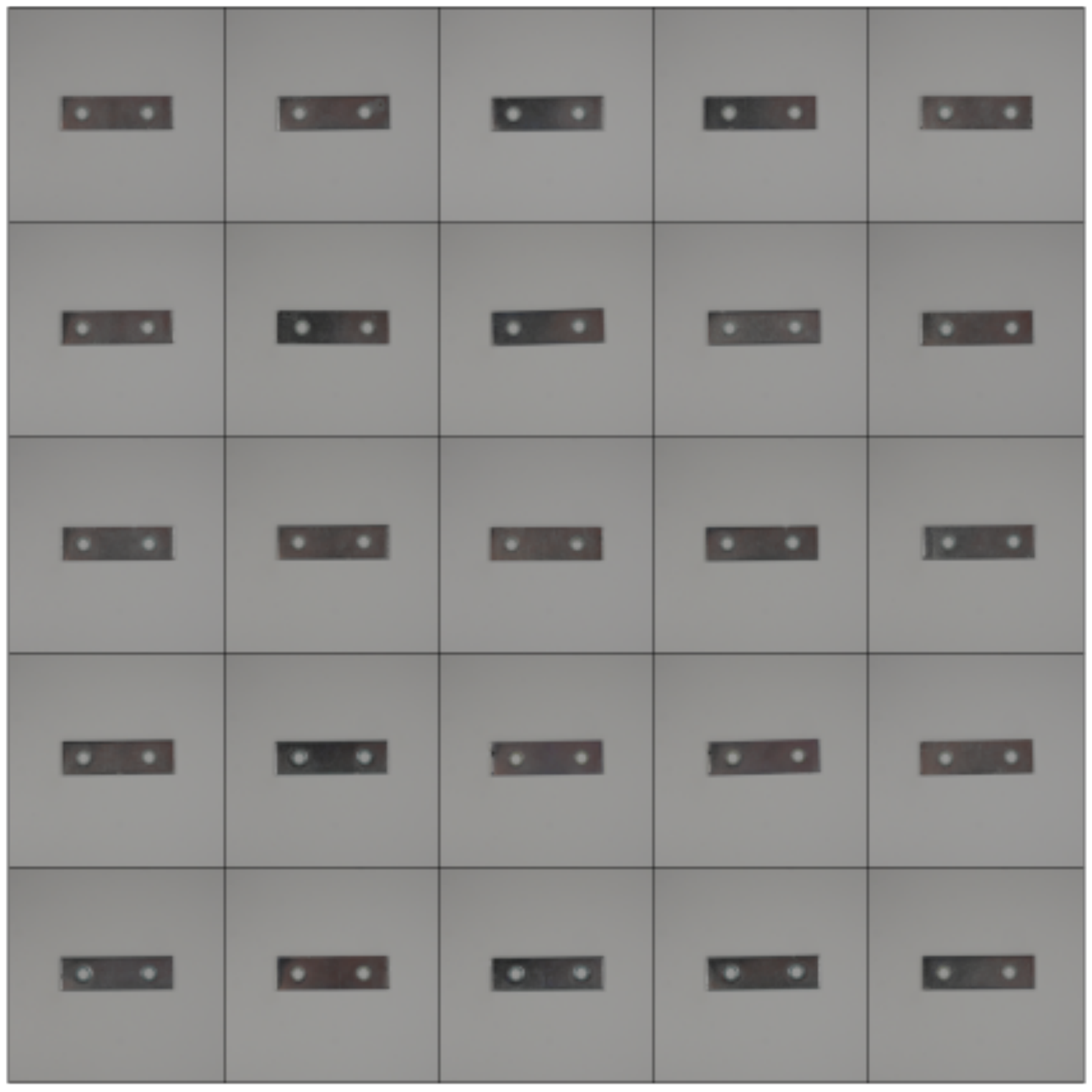}
	\caption{Mending plates data set.}
	\label{fig:Mendingplates1}
\end{figure}

\begin{figure}[H]
	\centering
	\includegraphics[width=5in]{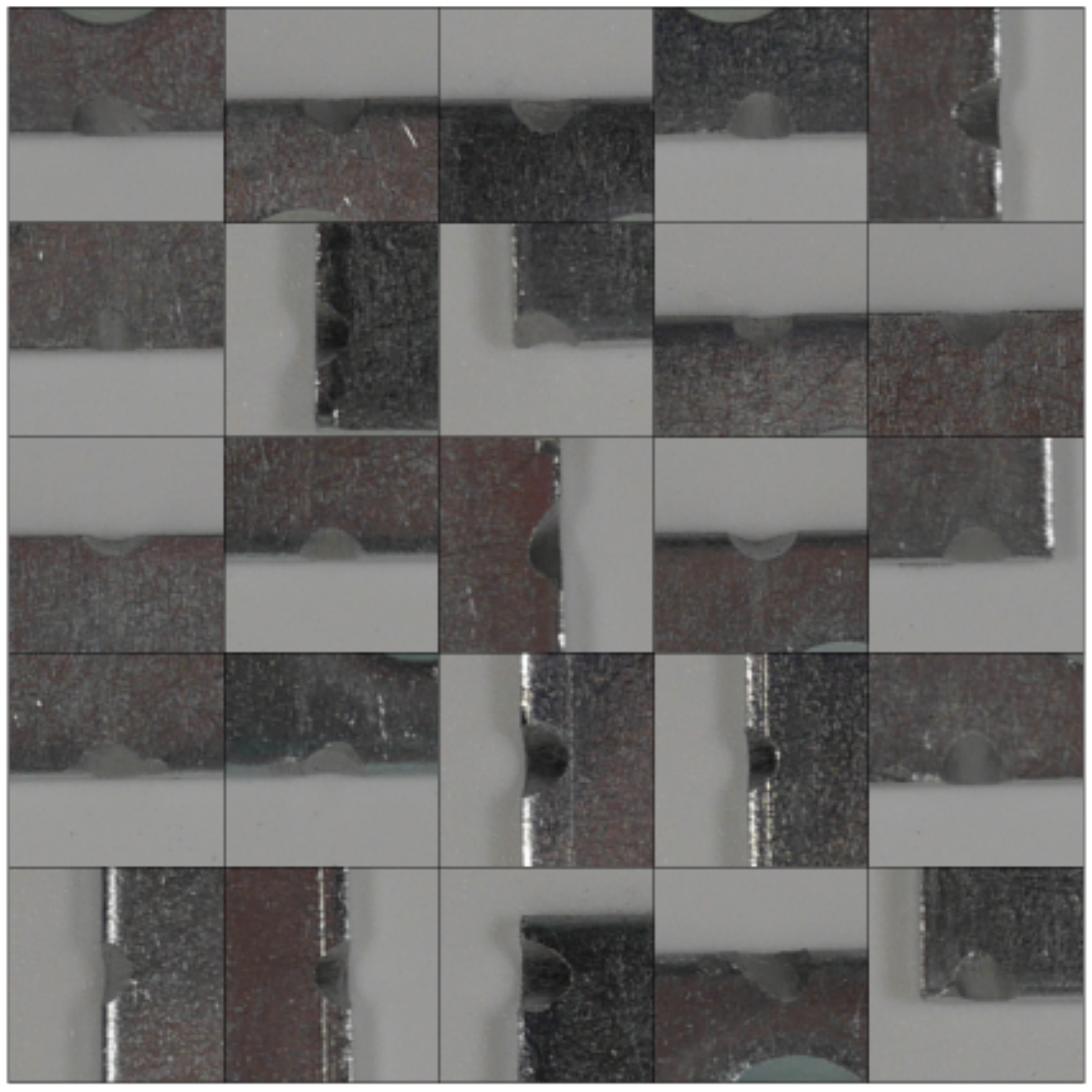}
	\caption{Zoomed areas showing defects introduced into a subset of the parts.}
	\label{fig:defects1}
\end{figure}

Fig. \ref{fig:defects1} shows close-up examples of the defects introduced. The defects are of a similar nature but vary in size, position, and orientation. The data consists of 160 \texttt{training} and \texttt{validation} images, consisting of parts that were purchased, had defects introduced, and were photographed, as one batch. Another 40, (“the \texttt{holdout} set”) were purchased, had defects introduced, and were photographed on a separate occasion and not involved in model training. 

Fig. \ref{fig:MendingplatesHO} below shows 8 example images of the Mending \texttt{holdout} set. They appear identical to the \texttt{train} and \texttt{validation} sets.

\begin{figure}[H]
	\centering
	\includegraphics[width=5in]{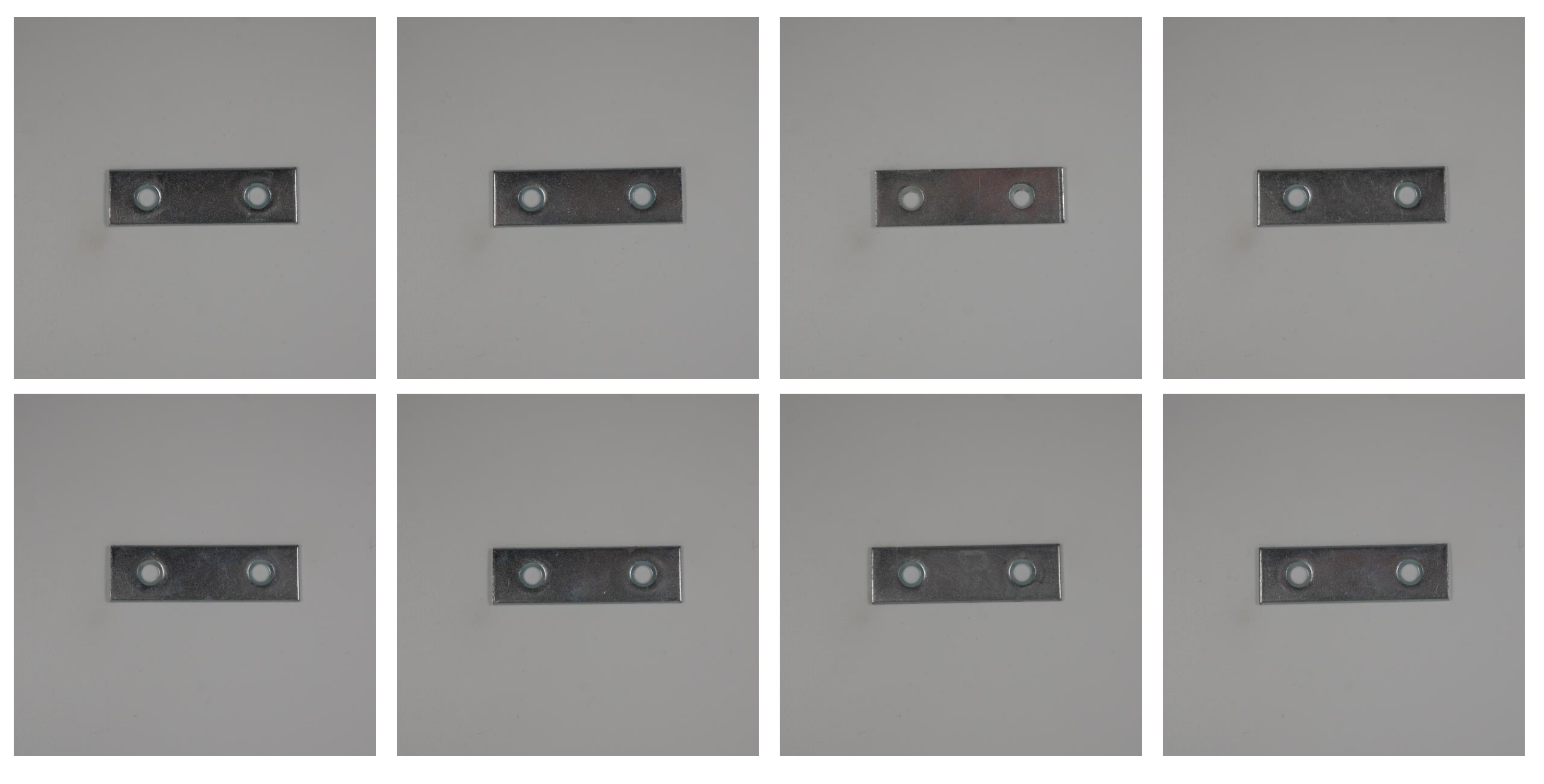}
	\caption{\texttt{Holdout} Mending plates data set.}
	\label{fig:MendingplatesHO}
\end{figure}

\begin{figure}[H]
	\centering
	\includegraphics[width=5in]{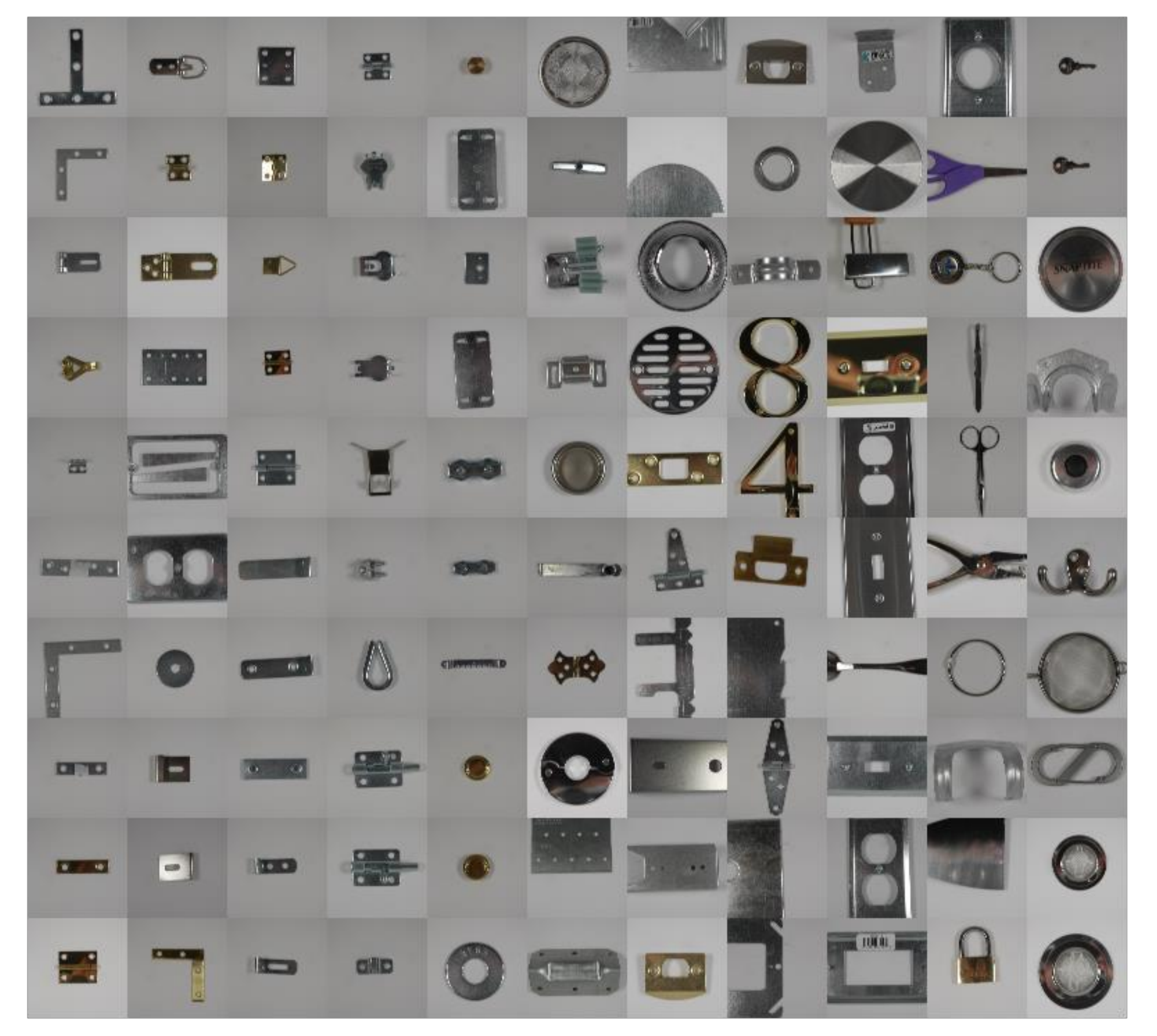}
	\caption{Generic data set.}
	\label{fig:underfill}
\end{figure}

A second data set consisting of photographs of duplicate copies of 132 flat metal parts was constructed, for 264 total. One of each part was damaged by removing a roughly crescent shape portion of material using a 3 mm diameter round file as with the “Mending plates”. The parts were photographed in four orientations each (undergoing planar rotations of nominally 0, 90 degrees, 180 degrees and 270 degrees).  Fig. \ref{fig:underfill} shows examples of the photographs. 220 of the parts were purchased, had defects introduced, and were photographed, as a batch, and comprise a \texttt{train} and \texttt{validation} set.

\begin{figure}[H]
	\centering
	\includegraphics[width=5in]{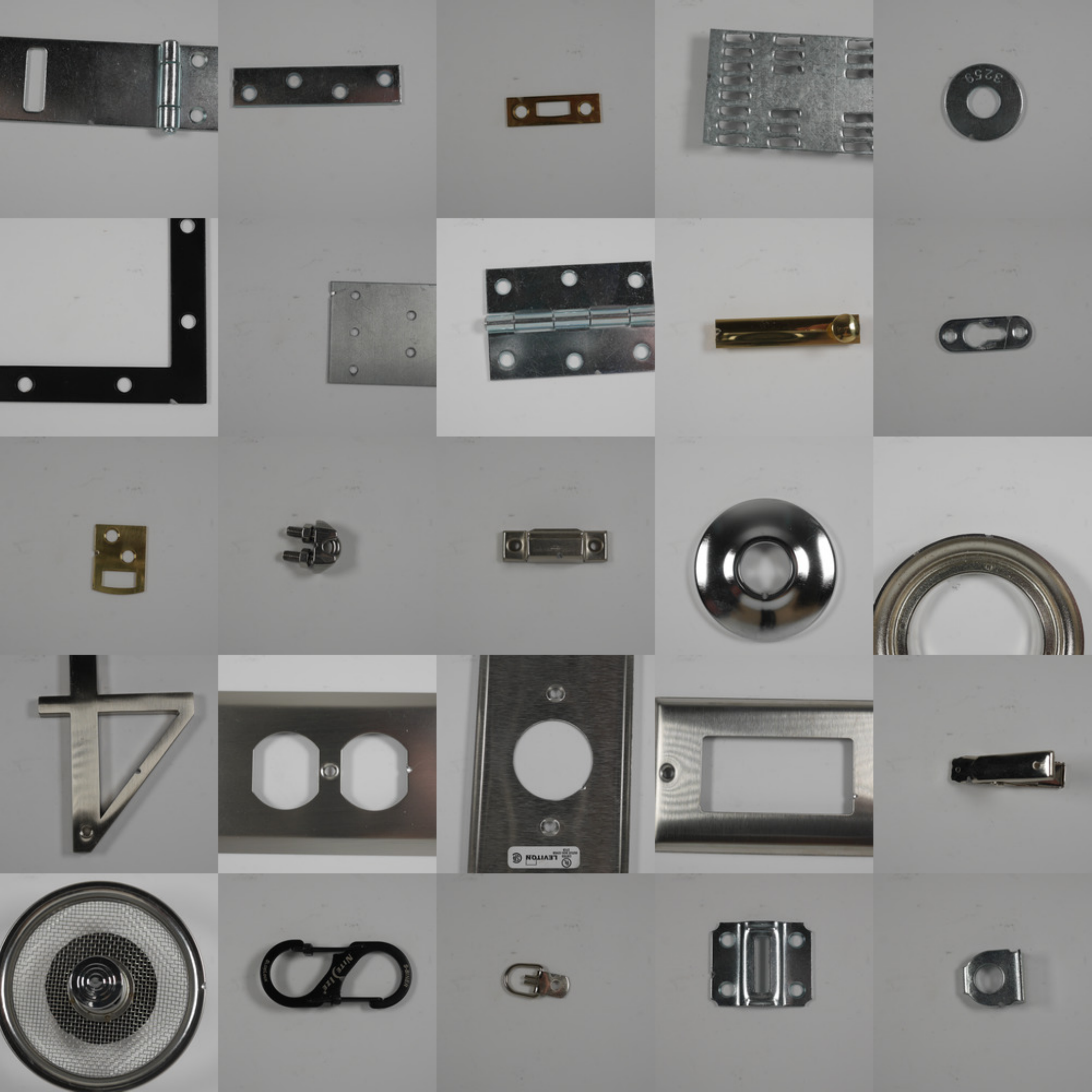}
	\caption{\texttt{Holdout} ``Generic'' data set.}
	\label{fig:underfillHO}
\end{figure}

The remaining 44 were purchased, had defects introduced, and photographed separately, and comprise a \texttt{holdout} set, as shown in Fig. \ref{fig:underfillHO}.

\begin{figure}[H]
	\centering
	\includegraphics[width=5in]{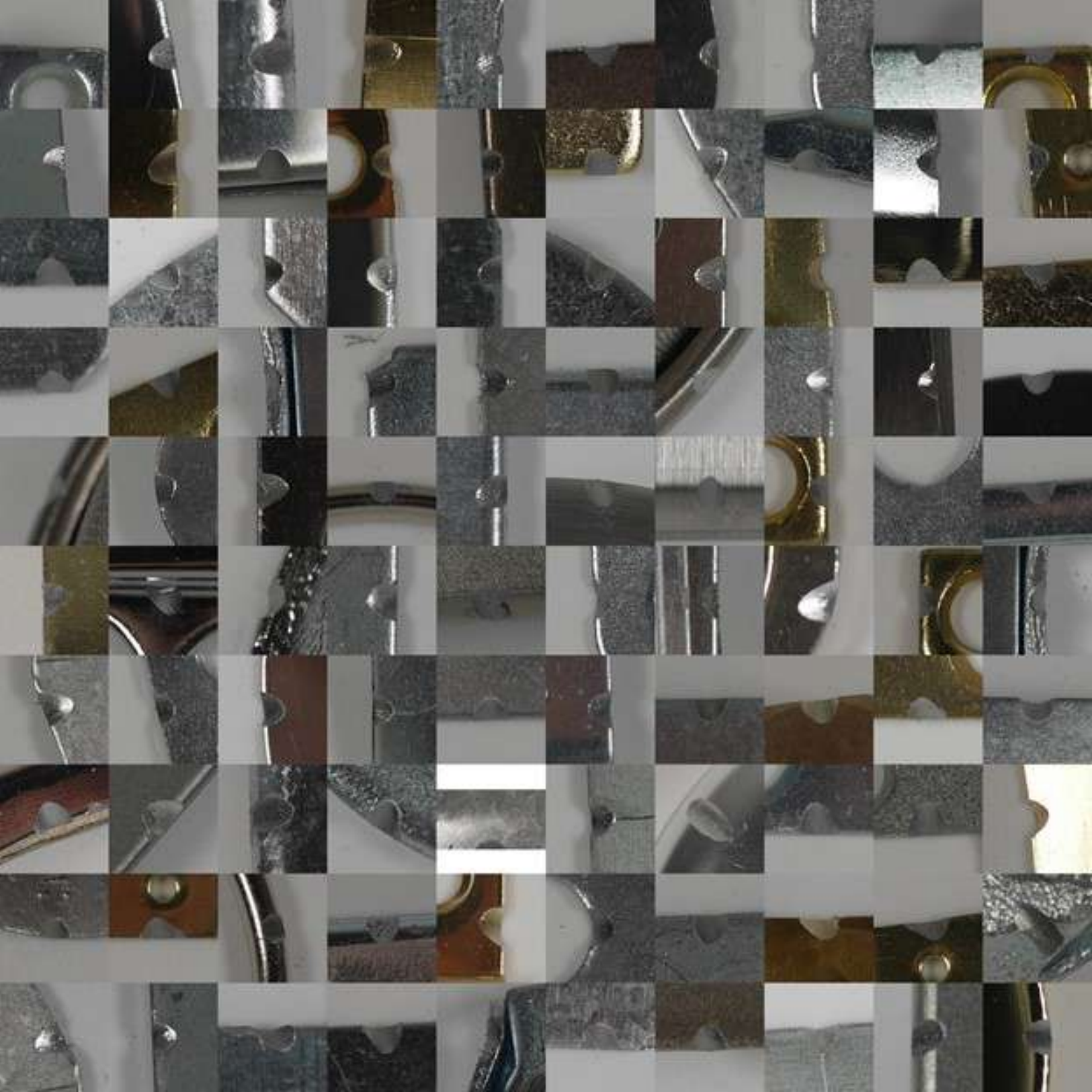}
	\caption{Zoomed areas showing defects introduced into a subset of the parts (``Generic'' data set).}
	\label{fig:defectsunder}
\end{figure}

Fig. \ref{fig:defectsunder} show a grid of examples of the defects introduced into the metal parts for the ``Generic'' datset. These defects are all of a similar type, but vary in expression based on the location, size, and other random variations.

The images in both the ``Mending plates'' and ``Generic'' data sets have a resolution of 2400×2400 pixels and were manually annotated with bounding boxes using labelstudio~\cite{LabelStudio}.





Table \ref{tab:com} provides a summary of the data set composition, including the names, original size, defect types, and whether weather augmentation was utilized.

\begin{table*}[!htpb]
\centering
\small
\caption{\label{tab:com}Data sets composition.}
\begin{tabular}{|l|l|l|l|}\toprule
\hline
\textbf{Data set} &  \textbf{Original data set size} &  \textbf{defects type} &  \textbf{Data Augmentation} \\ 
\midrule
\hline
“Mending plates”  & 140 & underfill and OKs & Yes\\ 
“Generic” & 110 & underfill and OKs & Yes \\
\hline
\bottomrule
\end{tabular}
\end{table*}






\section{Comparative results between a classifier and an object detection model}
\label{sec:res1}
\noindent
 A series of experiments with the ``Mending plates'' and ``Generic'' data sets was conducted to compare the performance of a classifier with that of an object detection model. The primary objective was to evaluate how effectively both models could learn to identify specific defects out of context and in held out examples, with the aim of improving their ability to detect such defects in new scenarios.

\subsection{Classifier}
\label{sec:res12}
\noindent
Table \ref{tab:class1} and Table \ref{tab:class2} provide the results for a set of experiments of the classifier for the two created data sets, the ``Mending plates'' and the ``Generic'' one.

For the ``Mending plates'' data set, the classifier performed almost perfectly over the \texttt{validation} set but failed over the \texttt{holdout} set of ``Mending plates''. The preparation and appearance of the \texttt{holdout} set is the same as for the \texttt{validation} and \texttt{training} set, except it was acquired on another day and not used in training. This simulates the case where a trained model is used in production. The poor performance suggests that the trained model is over-fit to spurious features in the training data, and despite performing well on the \texttt{validation} set (which is randomly split from the same pool as the training data) it is sufficiently fragile that it fails on seemingly identical data acquired later.

\begin{table*}[!htpb]
\centering
\small
\caption{\label{tab:class1}Set of experiments with the classifier.}
\begin{tabular}{|l|l|l|l|}\toprule
\hline
\textbf{Experiment} & \textbf{Train data} &  \textbf{Test data} &  \textbf{AUC} \\ \midrule
\hline
1 & ``Mending plates'' & \texttt{validation} set of ``Mending plates'' & 0.995\\
2 & ``Mending plates'' &  \texttt{holdout} set of ``Mending plates'' & 0.354  \\
\hline
\bottomrule
\end{tabular}
\end{table*}

Now, using the “Generic” data set, we can easily observe how the classifier performed well over the \texttt{validation} and the \texttt{holdout} set of the “Generic” data set. Compared with the “Mending plates”, the overall performance is lower (0.925 vs 0.995 on the \texttt{validation} set) but remains consistent to the \texttt{holdout} set. The key difference here is that the defects appeared in a diverse set of parts, forcing the model to focus on learning a set of features that was invariant to the background - i.e. the defect itself. Thus the model continues to perform even on ``new'' data acquired at a different time.

\begin{table*}[!htpb]
\centering
\small
\caption{\label{tab:class2}Set of experiments with the classifier.}
\begin{tabular}{|l|l|l|l|}\toprule
\hline
\textbf{Experiment} & \textbf{Train data} &  \textbf{Test data} &  \textbf{AUC} \\ \midrule
\hline
5 & ``Generic'' & \texttt{validation} set of ``Generic'' & 0.925\\
6 & ``Generic'' &  \texttt{holdout} set of ``Generic'' & 0.923  \\
\hline
\bottomrule
\end{tabular}
\end{table*}

\subsection{Object detection model}
\label{sec:res12}
\noindent

Table \ref{tab:objdm1} shows us that object detection model trained with the ``Mending plates'' data set is able to perform near perfectly over both the \texttt{validation} and the \texttt{holdout} set of ``Mending plates''. In both cases, the AUC is 0.99. Compared with the classifier, the object detection model is able to generalize to the \texttt{holdout} set, suggesting that the additional requirement to learn the bounding box location pushes the model to identify robust features that extend to new data, compared with the classifier.

\begin{table*}[!htpb]
\centering
\small
\caption{\label{tab:objdm1}Set of experiments with the object detection model.}
\begin{tabular}{|l|l|l|l|}\toprule
\hline
\textbf{Experiment} & \textbf{Train data} &  \textbf{Test data} &  \textbf{AUC} \\ \midrule
\hline
1 & ``Mending plates'' & \texttt{validation} set of ``Mending plates'' & 0.99\\
2 & ``Mending plates'' &  \texttt{holdout} set of ``Mending plates'' & 0.99  \\
\hline
\bottomrule
\end{tabular}
\end{table*}

Similar to the experiment conducted before, Table \ref{tab:objdm2} provides the results of object detection model trained on ``Generic'' data set and tested on the \texttt{validation} and \texttt{holdout} data set. The performance now approaches that for obtained on the more uniform “Mending plates” data, and generalizes to the \texttt{holdout} set.

\begin{table*}[!htpb]

\centering
\small
\caption{\label{tab:objdm2}Set of experiments with the object detection model.}
\begin{tabular}{|l|l|l|l|}\toprule
\hline
\textbf{Experiment} & \textbf{Train data} &  \textbf{Test data} &  \textbf{AUC} \\ \midrule
\hline
5 & ``Generic'' & \texttt{validation} set of ``Generic'' & 0.994\\
6 & ``Generic'' &  \texttt{holdout} set of ``Generic'' & 0.995  \\
\hline
\bottomrule
\end{tabular}
\end{table*}



\section{Generalization of the classifier and object detection models}
\label{sec:tl}

We use the term generalization to refer to a model's ability to learn general classification or detection rules that are robust to other changes in the data. Here we explore the generalization capabilities of a classifier and an object detection model on the ``Mending plates'' and ``Generic'' data sets. The aim is to investigate how well the models can generalize from one data set to another.

\subsection{Classifier}
\label{sec:MPD}
\noindent
Table \ref{tab:classTL} provides the results of the predictive power that the classifier has to generalize from "Mending plates" to the ``Generic'' data set and vice-versa. As observed from the results, the classifier failed to perform well (i.e., AUC equals to 0.5) going from from ``Mending plates'' to the ``Generic'' data set so no generalization was obtained in this case. However, it was able to generalize from the the ``Generic'' data set to the ``Mending plates'' data set with an AUC of almost 0.91 which is in keeping with the \texttt{validation} and \texttt{holdout} set results previously obtained for this classifier. We attribute this to the model's having learned a robust set of features for identifying defects. We conclude from this performance that it is actually better for in-production performance to train a classifier on a diverse set of data without examples of the part being inspected than to train on a uniform set of images of the same part. In practice, combining data sets to include diverse examples as well as real images of the part may be most appropriate.


\begin{table*}[!htpb]
\centering
\small
\caption{\label{tab:classTL}Set of experiments with the classifier.}
\begin{tabular}{|l|l|l|l|}\toprule
\hline
\textbf{Experiment} & \textbf{Train data} &  \textbf{Test data} &  \textbf{AUC} \\ \midrule
\hline
4 & ``Mending plates'' &  \texttt{holdout} set of ``Generic'' & 0.509   \\
7 & ``Generic'' &  \texttt{holdout} set of ``Mending plates'' & 0.911   \\
\hline
\bottomrule
\end{tabular}
\end{table*}



\subsection{Object detection model}
\label{sec:ODM22}
\noindent

Table \ref{tab:objdmTL} provides the results of the object detection model generalizing from ``Mending plates'' to the ``Generic'' data set and vice-versa. An object detection model trained on only the “Mending plates” is here able to achieve an AUC of 0.848 on the “Generic” data set. While this is lower than for a detector trained on the “Generic” train split, it shows a high degree of generalization to the defect type, despite only having seen examples of the defect in the “Mending plates”. This suggests the bounding box prediction requirement and associated training labels are causing the model to learn a robust set of features that generalize to defects in different circumstances, even from relatively uniform examples. When trained on the “Generic” data set and applied to the “Mending plates”, the model exhibits a perfect performance with an AUC of 1.0. This suggests that for a production model inspecting uniform parts, the combination of an object detection model and diverse training data can give the best performance.

\begin{table*}[!htpb]
\centering
\small
\caption{\label{tab:objdmTL}Set of experiments with the object detection model.}
\begin{tabular}{|l|l|l|l|}\toprule
\hline
\textbf{Experiment} & \textbf{Train data} &  \textbf{Test data} &  \textbf{AUC} \\ \midrule
\hline
4 & ``Mending plates'' &  \texttt{holdout} set of ``Generic'' &  0.848  \\
7 & ``Generic'' &  \texttt{holdout} set of ``Mending plates'' & 1.00   \\
\hline
\bottomrule
\end{tabular}
\end{table*}

\section{Factors that affect generalization}
\label{sec:res2}
\noindent




The main objective here is to identify the key factors that can affect the generalization of models, in order to create more representative data sets that focus on those significant factors. To achieve this, several steps are involved, beginning with image pre-processing, clustering, and finally determining the optimal number of clusters using a goodness measure. Clusters are identified using the \texttt{clustimage} package~\cite{clustimage} with agglomerative approach with the Euclidean distance metric and ward linkage. The Silhouette score~\cite{ROUSSEEUW198753} is used to assess both the clustering and the optimal number of clusters, with a search range between 3 and 25. The silhouette score ranges from -1 to 1, with higher values indicating better clustering. We chose the number of clusters that maximizes the silhouette score, which ensures that the clusters are well-separated and internally cohesive. Finally, the performance of the model on the testing data set will be evaluated using different training data sets generated through clustering. This approach will produce multiple training data sets for subsequent experiments, each with unique characteristics and potentially different results.

\subsection{Clustering on the ``Generic'' data set}
\label{sec:c_generic}
\noindent

We analyzed 110 images from the “Generic” data set by clustering them and then assessed the level of generalization by testing the resulting models on the ``Mending plates'' data set. Using the ``silhouette''~\cite{ROUSSEEUW198753} evaluation method, we determined that the best number of clusters was three, which is illustrated below in Figs. \ref{fig:c0}, \ref{fig:c1}, and \ref{fig:c2}, respectively. We can see that the clusters roughly represent different shapes that are present in the data set. Cluster one is the most diverse and includes smaller items with more white space around them. Cluster 2 includes bigger items and in particular some with large holes such as the metal ``8'' and outlet covers. And Cluster 3 contains many larger round objects.


We produced three separate training data sets by excluding the images that are present in each cluster. This way, we trained models with the remaining images available in every created training data set. 

\begin{figure}[H]
	\centering
	\includegraphics[width=5in]{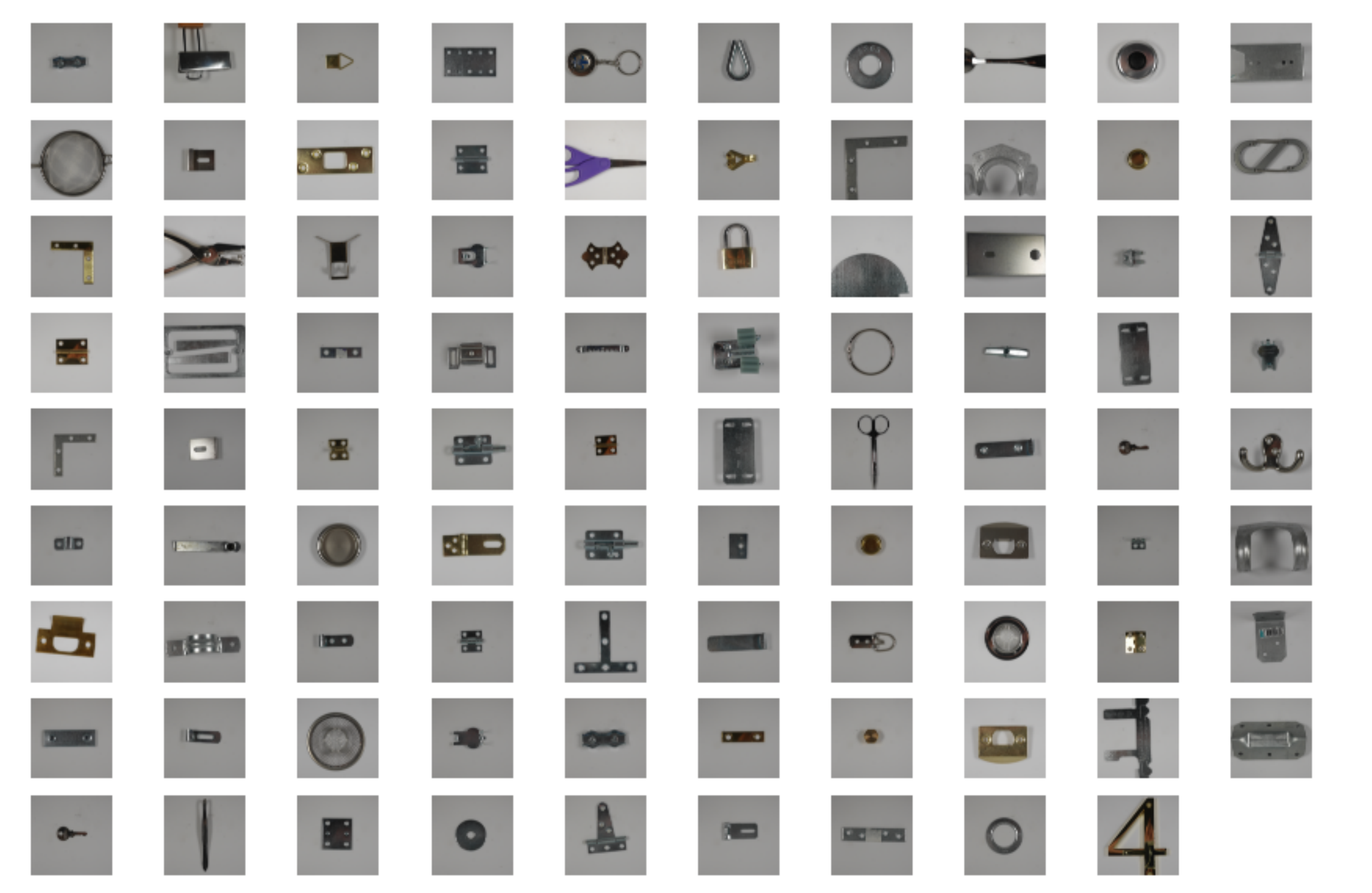}
	\caption{Cluster 1: images with varying subjects and compositions, featuring smaller items and more negative space.}
	\label{fig:c0}
\end{figure}

\begin{figure}[H]
	\centering
	\includegraphics[width=5in]{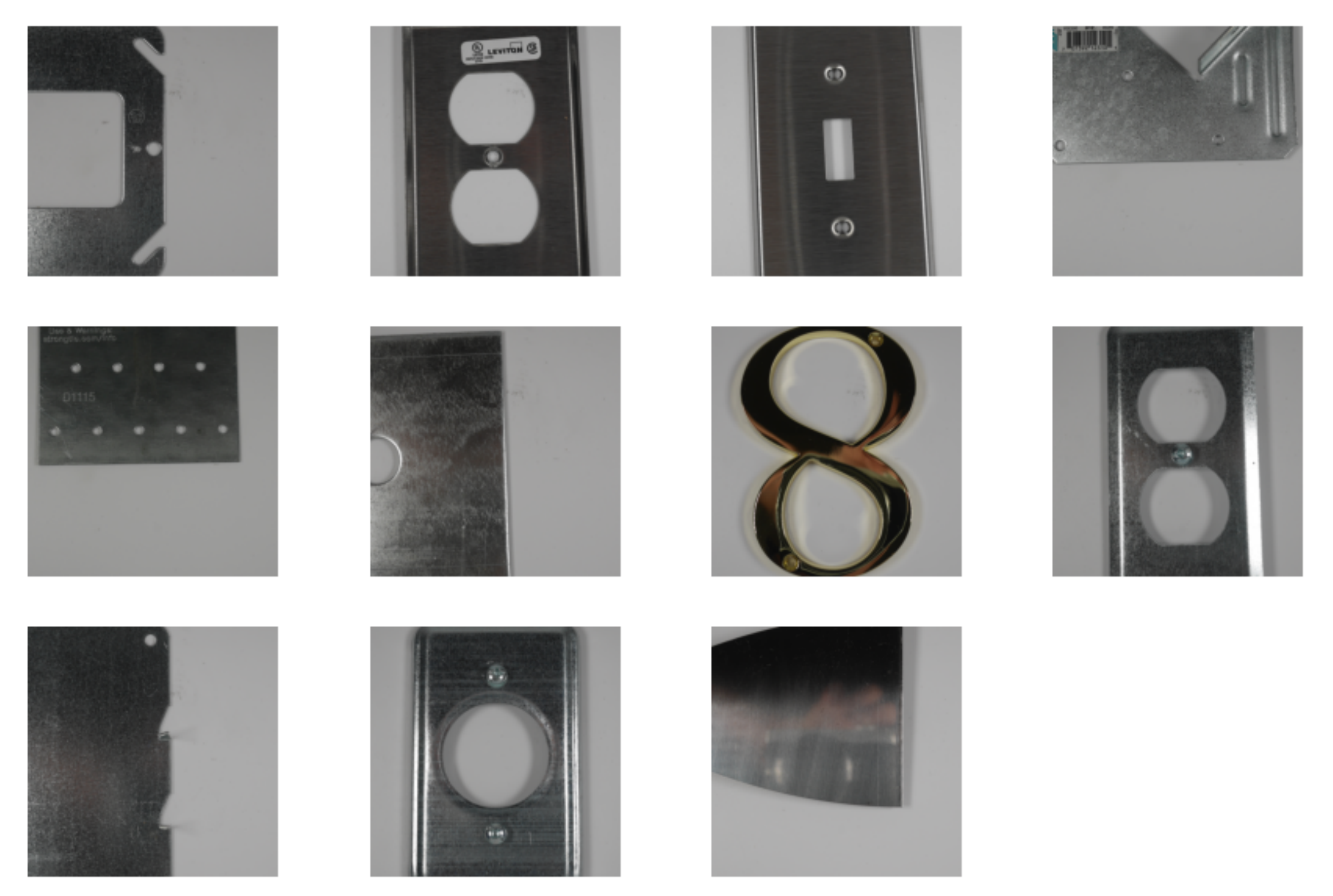}
	\caption{Cluster 2: images with larger, perforated items such as metal numbers and outlet covers.}
	\label{fig:c1}
\end{figure}

\begin{figure}[H]
	\centering
	\includegraphics[width=5in]{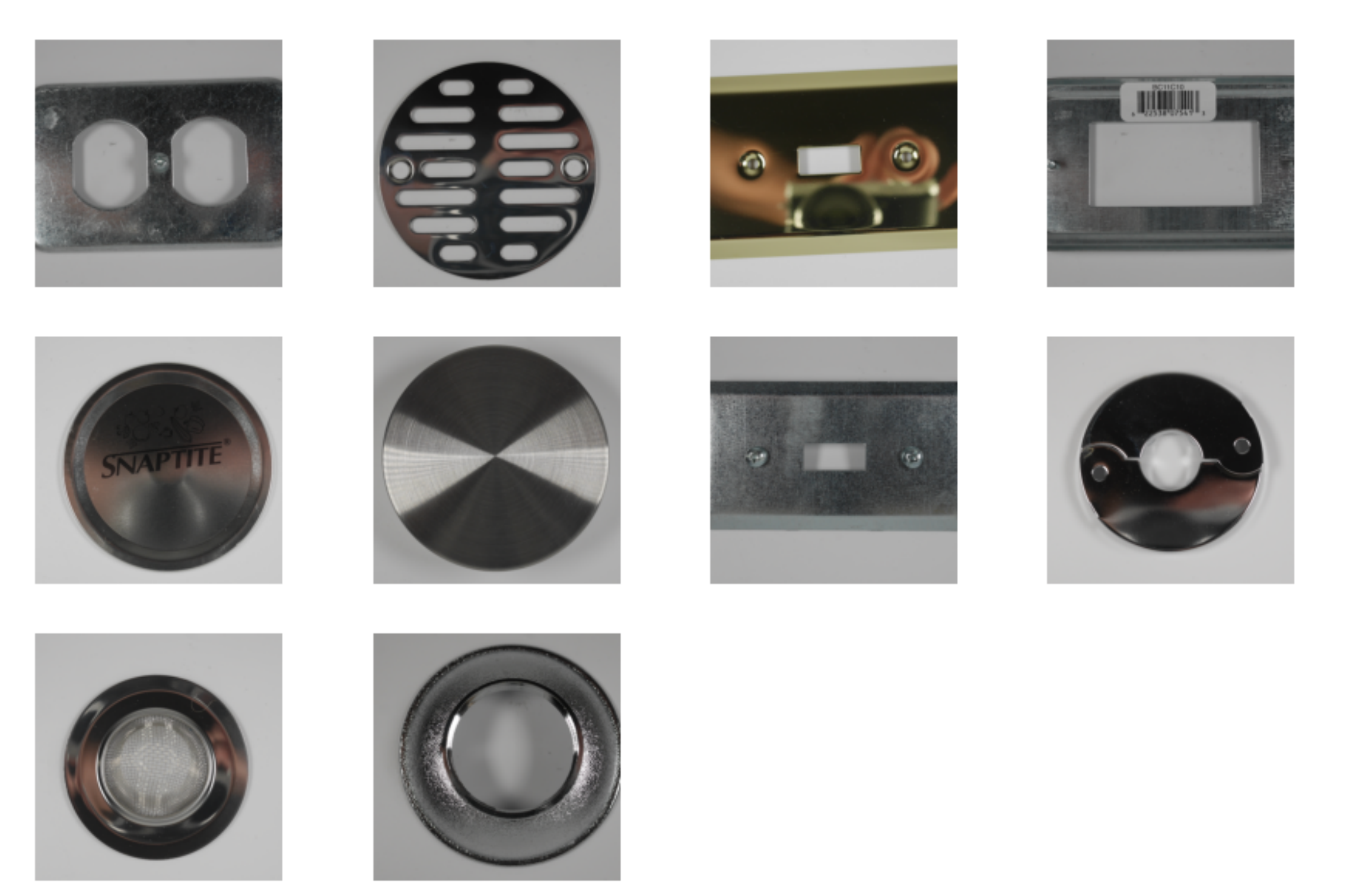}
	\caption{Cluster 3: images featuring large circular objects, with some variations in size.}
	\label{fig:c2}
\end{figure}

The results obtained before and after utilizing our approach are presented in Table  \ref{tab:uti}. Experiment 0, which did not employ any clustering, included all images in the training data set. In Experiment 1, images belonging to cluster 1 were excluded from the training data set. In Experiment 2, images belonging to cluster 3 were eliminated, and in Experiment 3, images belonging to cluster 3 were omitted. These exclusions represented 80\%, 10\%, and 9\% of the entire training data set, respectively.


Omitting cluster 1 in Experiment 1 has the largest impact, dropping the performance from and AUC of 0.995 to 0.923. However, even with 80\% of the data removed, this performance drop only amounts to about 7\%.

The results of Experiment 2 indicates that cluster 2 has no impact on the generalization as the AUC maintained equal compared to the case of not employing any clustering method.

Finally, the results of experiment 3 indicates that cluster 3 contributes to the overall performance we have seen a drop in the AUC value compared to the case of not employing any clustering method, from 0.995 to 0.985.

Overall these results suggest that we can eliminate some images from the training data without decreasing performance, and that it may be possible to improve performance by selectively including additional data in clusters tho which the performance is sensitive. A higher number of clusters could be used to fine-tune recommendations for adding or omitting data.

\begin{table*}[!htpb]
\centering
\small
\caption{\label{tab:uti}Set of experiments over the obtained clusters.}
\begin{tabular}{|l|l|l|}\toprule
\hline
\textbf{Number of experiment} & \textbf{number of images excluded} &  \textbf{AUC on Test data} \\ \midrule
\hline
0 & 0 & 0.995\\
1  & 89 & 0.923\\
2 &  11& 0.995  \\
3 & 10 &   0.985 \\
\hline
\bottomrule
\end{tabular}
\end{table*}



\section{Conclusion and future work}
\label{sec:Conclusion}
\noindent
 
We have examined how data diversity contributes to model robustness classification and object detection models, in a typical manufacturing inspection context. When trained on repetitive data, binary OK/NG classifiers are brittle and may not even generalize to seemingly identical held out data, as demonstrated by our experiments on “Mending plates” images. A classifier can be made more robust by training on diverse data where a defect is presented on different backgrounds. In our experiments, we found that training on similar defects in diverse images of flat metal parts results in a ~0.92 classifier AUC on a validation set that is maintained when predicting on a held-out data set. We conclude that the diverse data forces the model to learn an invariant set of characteristics of the training data that generalizes to new image. We shows that a classifier trained on diverse data performs equally well on a uniform data set. Put together, this provides a recipe for training a model with good in-distribution performance that is known to be robust to defects in new contexts.

By moving from a classifier to an object detection model, we can further improve performance. The additional requirement to localize the defect acts as a constraint that forces the model to learn a robust set of features. Even a model trained on a highly uniform data set has been shown to generalize well to a diverse data set (AUC of ~0.85). Combining object detection with a diverse training data set yields the best performance.

We used a clustering method to examine which data was most important to the model performance. This is important to minimize the effort required in data collection and labeling. Certain clusters were found to have a lower importance for performance, which can inform data collection.

This study focused on a single defect type, representing missing material. The background part was varied to examine how well the model generalizes. In ongoing work we are studying how well defects generalize between different types. The overall goal is a recipe for data collection and model validation that ensures robust performance that can be sustained on newly encountered defects.

\section{Acknowledgment}
\noindent

Ahmad Mohamad Mezher is a recipient of a McCain Postdoctoral Fellowship in Innovation with the Electrical and Computer Engineering department at the University of New Brunswick (UNB).

\bibliographystyle{IEEEtranS}
\bibliography{MarbleMezher}

\end{document}